\documentclass[12pt]{article}
\usepackage[utf8]{inputenc}
\usepackage{authblk}

\title{Information Potential Auto-Encoders}

\author{Yan Zhang, Mete Ozay, Zhun Sun and Takayuki Okatani 
}
\affil{Graduate School of Information Sciences, 
Tohoku University}
\affil{Aramaki Aza Aoba, Aoba-ku, 980-8579 Sendai, Japan}
\affil{\{zhang, mozay, sun, okatani\}@vision.is.tohoku.ac.jp}

\date{}

\usepackage[utf8]{inputenc} % allow utf-8 input
\usepackage[T1]{fontenc}    % use 8-bit T1 fonts
\usepackage{url}            % simple URL typesetting
\usepackage{booktabs}       % professional-quality tables
\usepackage{amsfonts}       % blackboard math symbols
\usepackage{nicefrac}       % compact symbols for 1/2, etc.
\usepackage{microtype}      % microtypography
\usepackage{algorithm}
\usepackage{algorithmic}
\usepackage{mathtools}
\usepackage{amssymb}
\usepackage{bbm}
\usepackage{epsfig}
\usepackage{graphicx}
\usepackage{amsmath}
\usepackage{amssymb}
\usepackage{times}
\usepackage{epsfig}
\usepackage{graphicx}
\usepackage{amsmath}
\usepackage{amssymb}
\usepackage{subcaption}
\usepackage{makecell}
\usepackage{amsthm}
\usepackage{soul}
\usepackage{multirow}
\usepackage{bm}

\theoremstyle{definition}

\theoremstyle{definition}

\theoremstyle{remark}

\theoremstyle{remark}

\theoremstyle{remark}

\makeatletter

\newcommand{\xdashrightarrow}[2][]{\ext@arrow 0359\rightarrowfill@@{#1}{#2}}
\newcommand{\xdashleftarrow}[2][]{\ext@arrow 3095\leftarrowfill@@{#1}{#2}}
\newcommand{\xdashleftrightarrow}[2][]{\ext@arrow 3359\leftrightarrowfill@@{#1}{#2}}
\def\rightarrowfill@@{\arrowfill@@\relax\relbar\rightarrow}
\def\leftarrowfill@@{\arrowfill@@\leftarrow\relbar\relax}
\def\leftrightarrowfill@@{\arrowfill@@\leftarrow\relbar\rightarrow}
\def\arrowfill@@#1#2#3#4{%
  $\m@th\thickmuskip0mu\medmuskip\thickmuskip\thinmuskip\thickmuskip
   \relax#4#1
   \xleaders\hbox{$#4#2$}\hfill
   #3$%
}

\begin{document}

\maketitle

\begin{abstract}
In this paper, we suggest a framework to make use of mutual information as a regularization criterion to train Auto-Encoders (AEs). In the proposed framework, AEs are regularized by minimization of the mutual information between input and encoding variables of AEs during the training phase. In order to estimate the entropy of the encoding variables and the mutual information, we propose a non-parametric method. 
We also give an information theoretic view of Variational AEs (VAEs), which suggests that VAEs can be considered as parametric methods that estimate entropy. Experimental results show that the proposed non-parametric models have more degree of freedom in terms of representation learning of features drawn from complex distributions such as Mixture of Gaussians, compared to methods which estimate entropy using parametric approaches, such as Variational AEs. 

\end{abstract}

\section{Introduction}
% Recently, deep networks are widely used in unsupervised learning tasks. 

We address a \textit{representation learning} problem as follows.
Let $\mathbf{x}$ be the input (observed) variable. In our problem of interest, we consider the target of the representation learning as learning an encoding function $f(\cdot)$, where the input variable
$\mathbf{x}$ is encoded to a new variable, $\mathbf{z}=f(\mathbf{x})$, called as an encoding variable.  

Auto-encoders (AEs) are popular unsupervised representation learning methods, in which the encoding functions are implemented by neural networks, i.e. encoders. The encoders are coupled by decoders which aim to reconstruct the input variable from the encode variable.
AEs are learned by minimizing the reconstruction error.
Using this learning scheme, the encoder is expected to capture the underlying \textit{useful} structure of the input variable. However, a trivial solution to such an optimization problem can be obtained by setting an identity mapping between the input variable and the encode variable. The learned encoder cannot be useful with a trivial solution.
Thus, regularization methods are required to avoid trivial solutions and learn representations using AEs.

% Recently, deep networks are widely used in unsupervised learning tasks. 
% Among the unsupervised deep learning methods, the Auto-Encoder family plays an important role. 
% A common architecture of Auto-Encoders usually consists of two parts. 
% The \textit{encoder} network transforms an input variable to a new variable, notated as \textit{encoding variable} in this paper. T
% he \textit{decoder} network aims to reconstruct the input variable from the encoding variable. 

In this work, we consider development of regularization methods for AEs from an information theoretic perspective.
 Information theoretic approaches have been widely used to develop regularizers or objectives utilized for learning of representations \cite{ib,eminon,eminon2}.
Mutual information $I(\mathbf{x};\mathbf{z})$  between input variable $\mathbf{x}$ and encoding variable $\mathbf{z}$ has been employed as a regularizer for representation learning. $I(\mathbf{x};\mathbf{z})$ quantifies the information content between $\mathbf{x}$ and $\mathbf{z}$. If we consider the encoding function as a compressor which compresses $\mathbf{x}$ into $\mathbf{z}$, then 
$I(\mathbf{x};\mathbf{z})$ can be seen as a measure of compression degree. Theoretical results given by \cite{ibgene} indicate that a boost of generalizing performance can be obtained by regularizing supervised learning methods using $I(\mathbf{x};\mathbf{z})$ in supervised learning problems. In unsupervised learning tasks, mutual information has been also used to develop regularizers as shown by \cite{eminon}, and \cite{eminon2}. 

Estimation of mutual information $I(\mathbf{x};\mathbf{z})$ is a key task for employment of information theoretic regularizers. Obviously, $I(\mathbf{x};\mathbf{z})$ can be decomposed into two entropy terms, i) $H(\mathbf{z}|\mathbf{x})$, and ii) $H(\mathbf{z})$. The first term requires modeling of a conditional distribution $p(\mathbf{z}|\mathbf{x})$. Recently, a stochastic encoding mechanism has been introduced to deep networks by \cite{vae}.
In this encoding mechanism, we can obtain the distribution $p(\mathbf{z}|\mathbf{x})$, and thus
$H(\mathbf{z}|\mathbf{x})$ can be analytically computed. Meanwhile the stochastic sampling from $p(\mathbf{z}|\mathbf{x})$ during the backpropagation can be efficiently performed by a re-parameterization trick. Thereby, $H(\mathbf{z}|\mathbf{x})$ is computed using a parametric model.
As for $H(\mathbf{z})$, we need to estimate $p(\mathbf{z})$. A solution of this challenge is employment of a variational distribution $q(\mathbf{z})$ to obtain an upper bound of $H(\mathbf{z})$ as shown by \cite{vibg}. However, in order to use the variational distribution, it is assumed that $p(\mathbf{z})$ has the same distribution as $q(\mathbf{z})$, which is a pre-defined distribution such as Gaussian. Thus, $H(\mathbf{z})$ is estimated under a parameteric model. In many cases, we cannot assure that an encoding variable should be drawn from a Gaussian distribution.

To this end, we propose a non-parametric method in order to compute $H(\mathbf{z})$. Our contributions can be summarized as follows.
\begin{enumerate}
    \item In our proposed method, it is not required to make any assumption for a particular parameterized form of distribution $p(\mathbf{z})$ of an encoding variable. This property gives more degree of freedom to learn the distribution of $p(\mathbf{z})$ during the representation learning tasks, as examined in the experimental analyses. 
    
    \item We employed the proposed method to train AEs with mutual information regularization. As shown in Section 2, by the employment of the proposed method, regularization of AEs by minimization of mutual information can be considered as minimization of the pair-wise distance between samples in the encoding space. Therefore, we obtain an analogue to the \textit{information potentials} scheme of physical particles \cite{eminon} in training of AEs.
The information potential scheme of AEs aims to map any two different samples from the input space to the same sample in the encoding space. It enables AEs to avoid trivial solutions of identity mapping. We call AEs trained by the proposed method, Information Potential AEs (IPAEs).

\item We provide an information theoretic view of Variational Auto-Encoders (VAEs). The regularization term employed by VAEs has the same form of minimization of mutual information used in parametric methods. We show that the mutual information regularization used in VAEs can be considered as %reduction of the distances between all samples 
mapping of all samples to a point (e.g. $\mathbf{0}$) in the encoding space. On the other hand, the proposed method minimizes the mutual information by reducing pair-wise distances. Experimental results show that our method proposed for regularization of AEs by minimization of mutual information enables IPAEs to have more degree of freedom in terms of representation learning of features drawn from complex distributions such as Mixture of Gaussians, compared to VAEs. 
% Additionally, it is shown in the experiments that , without computing all pair of distance, IPAEs can also obtain 
% can be trained as fast as VAEs even with non-parametric modeling.

\end{enumerate}

%In other words, our method is not restricted to some specific encode distribution, it can be arbitrary.

% Non-parametric methods have been investigated in \cite{eminon,eminon2}.

% stochastic encoding mechanism has been employed as a regularization method for learning deep representations in some previous works \cite{vae,vibg,vibuc}.

% Specifically, we adopt a stochastic encoding scheme in which the encoding variable given an input variable follows a conditional distribution. Instead of directly minimizing the reconstruction error, the optimization objective of the proposed framework is to minimizing the Mutual Information between the input variable and the encoding variable under the constraint that the reconstruction error is smaller than a given threshold. Efficient estimation of mutual information is critical in the proposed system. The difficulty is to compute the Entropy of encoding variable.
% In previous works such as \cite{vibg}, the entropy of encoding variable is approximated by introducing an variational distribution. However, the introduced variational distributions are usually parametric distributions. In this work, we propose a non-parametric method for estimating it. The propose method can be efficiently incorporated into learning system of deep networks.

The rest part of this paper is organized as follows. In Section 2, we introduce our proposed methods. In Section 3, we review the related work. Section 4 illustrates experimental analyses and results. Finally, the conclusion is given in Section 5.

 \section{Methods}
 \subsection{Problem Definition}
Let a random variable $\mathbf{x}$ be the input variable to an AE, and $f(\cdot)$ and $g(\cdot)$ be mapping functions of an encoder and a decoder, respectively. Encoding variable is denoted by $\mathbf{z}:=f(\mathbf{x})$. 
% Let $\tilde{\mathbf{x}}$ be the reconstructed representation: $\tilde{\mathbf{x}}=g(\mathbf{z})$. 
Suppose that we are given a set of i.i.d. observations $\mathcal{X}=\{\mathbf{x}_i \}_{i=1}^N$. Then, our framework proposed for training of AEs is as follows.
We define an optimization objective by minimizing the mutual information between the input variable and the encoding variable under the constraint that the reconstruction error is smaller than a given threshold, which is formulated by
\begin{equation}
\begin{split} 
&\min_{f,g} I(\mathbf{x};\mathbf{z}) \\
&\text{s.t.}\quad \mathbb{E}_{\mathbf{x}\sim p(\mathbf{x})} [d(\textbf{x},g(\mathbf{z}))] \leq D
 \end{split}
 \label{eq:rd}
\end{equation}
where $d(\cdot,\cdot)$ is a distance function, and $D$ is a given threshold. The objective can be seen as a rate distortion problem, which aims to find the minimal number of bits (measured by $I(\mathbf{x};\mathbf{z})$) required to encode $\mathbf{x}$ into $\mathbf{z}$ so that the source $\mathbf{x}$ can be reconstructed without exceeding a given distortion threshold $D$, with respect to the reconstruction error $d(\textbf{x},g(\mathbf{z}))$. The rate distortion theory has been used to learn AEs by \cite{rateae}. The difference comes from the mutual information term; they aim to minimize $I(\mathbf{x};g(\mathbf{z}))$ while our goal is to minimize $I(\mathbf{x};\mathbf{z})$.

Our encoder can be considered as a lossy data compression model, where the degree of compression is quantified by the information content of the relationship between an input variable and the corresponding encoding variable. A large distortion threshold will give more freedom to AEs to compress the input variable into a smaller number of bits.
The criteria of compressing information content will drive the encoder to extract \textit{useful} information from $\mathbf{x}$, and to discard noisy part, and thus avoid trivial solutions.

By introducing a Lagrange multiplier, \eqref{eq:rd} can be reformulated by
\begin{equation}
\begin{split} 
&\min_{f,g}\ \mathbb{E}_{\mathbf{x}\sim p(\mathbf{x})} [d(\textbf{x},g(\mathbf{z}))] + \beta\, (H(\mathbf{z})-H(\mathbf{z}|\mathbf{x}) ).
 \end{split}
 \label{eq:rd2}
\end{equation}
where $\beta$ is the inverse value of the introduced Lagrange multiplier controlling a trade-off between mutual information loss and reconstruction loss.
In this work, we adopt a stochastic encoding scheme for the encoders, in which the backpropagation of a stochastic encoder can be efficiently performed by a re-parameterization trick as shown in \cite{vae}. More precisely, the encoder mapping function is formulated by a Gaussian stochastic mapping
\begin{equation}
\begin{split} 
\mathbf{z}&=f(\mathbf{x}),\\
&=\mu(\mathbf{x})+\sigma(\mathbf{x})\odot\bm{\epsilon},
 \end{split}
 \label{eq:pzx}
\end{equation}
where $\odot$ denotes element wise product, $\mu (\cdot)$ and $\sigma (\cdot)$ are computed by encoder networks, and $\bm{\epsilon}\sim \mathcal{N}(\mathbf{0},\mathbf{I})$. As a result, the conditional distribution of $\mathbf{z}$ given an input variable follows a Gaussian distribution $p(\mathbf{z}|\mathbf{x})=\mathcal{N}(\mu(\mathbf{x}),\text{diag}(\sigma^2(\mathbf{x})))$. In this case, conditional entropy $H(\mathbf{z}|\mathbf{x})$ can be analytically computed. Next, we consider the problem of estimating $H(\mathbf{z})$.

\subsection{Estimation of $H(\mathbf{z})$ via Parametric Ways}

We first consider parametric methods for estimating $H(\mathbf{z})$.
One approach employed to solve this problem is to first identify a variational distribution $q(\mathbf{z})$. Then, its upper bound is obtained by using the property that Kullback-Leibler divergence is always positive, i.e.
\begin{equation*}
\begin{split}
& KL(p(\mathbf{z})||q(\mathbf{z}))\geq 0\ \\
\Rightarrow &\ \sum_{\mathbf{z}} p(\mathbf{z}) \log p(\mathbf{z}) \geq \sum_{\mathbf{z}} p(\mathbf{z}) \log q(\mathbf{z}) \\
\Rightarrow & H(\mathbf{z}) \leq - \sum_{\mathbf{z}} p(\mathbf{z}) \log q(\mathbf{z}).
 \end{split}
\end{equation*}
In this case, the distribution of $q(\mathbf{z})$ can be assumed to be a pre-defined distribution, e.g. Gaussian $\mathcal{N}(\mathbf{0},\mathbf{I})$, as suggested by \cite{vibg}. As a result, the upper bound of mutual information $I(\mathbf{x};\mathbf{z})$ can be computed by
\begin{equation}
\begin{split} 
I(\mathbf{x};\mathbf{z})&=H(\mathbf{z})-H(\mathbf{z}|\mathbf{x}),\\
&=\mathbb{E}_{\mathbf{z},\mathbf{x}\sim p(\mathbf{z},\mathbf{x})}[-\log p(\mathbf{z})+\log p(\mathbf{z}|\mathbf{x})]\\
&\leq \mathbb{E}_{\mathbf{z},\mathbf{x}\sim p(\mathbf{z},\mathbf{x})}[-\log q(\mathbf{z})+\log p(\mathbf{z}|\mathbf{x})].
 \end{split}
 \label{eq:mi1}
\end{equation}
% where $H(\cdot)$ denotes entropy. 
Note that this bound is exactly the same as the regularization term used by Variational Auto-Encoders (VAEs).

In \eqref{eq:mi1}, we first replace the expectation operation with summation of empirical samples and $p(\mathbf{z}|\mathbf{x})=\mathcal{N}(\mu(\mathbf{x}),\text{diag}(\sigma^2(\mathbf{x})))$, $q(\mathbf{z})=\mathcal{N}(\mathbf{0},\mathbf{I})$. Then, we obtain an empirical formulation of the upper bound of the mutual information using a parametric method,
\begin{equation}
\begin{split} 
I(\mathbf{x};\mathbf{z})\leq \frac{1}{2N} \sum \limits_{i=1}^N \Big(\big|\big| \mu(\mathbf{x}_i)  \big|\big|_2^2 + \big|\big| \sigma^2(\mathbf{x}_i)\big|\big|_1 - \log \big| \text{diag}(\sigma^2(\mathbf{x}_i)) \big|-1   \Big),
 \end{split}
 \label{eq:mi2}
\end{equation}
where $||\cdot||_2^2$ is the square of the $\ell_2$ norm, $||\cdot||_1$ denotes $\ell_1$ norm and $|\cdot|$ denotes the matrix determinant. 
% Applying \eqref{eq:mi2} into objective 
% I(\mathbf{x};\mathbf{z})\simeq \frac{1}{N} \sum \limits_{i=1}^N \mathbb{E}_{\mathbf{z}\sim p(\mathbf{z}|\mathbf{x}_i)}\big[-\log  p(\mathbf{z}) + \log p(\mathbf{z}|\mathbf{x}_i) \big].

\subsection{Our Proposed Non-parametric Method}

% . In \eqref{eq:mi2}, we consider the problem of estimating $\log p(\mathbf{z})$. One approach employed to solve this problem is to first identify a variational distribution $q(\mathbf{z})$. Then, we obtain its upper bound using the property that Kullback-Leibler divergence is always positive, i.e.
% \[
% KL(p(\mathbf{z})||q(\mathbf{z}))\geq 0\ \Rightarrow \ \sum_{\mathbf{z}} p(\mathbf{z}) \log p(\mathbf{z}) \geq \sum_{\mathbf{z}} p(\mathbf{z}) \log q(\mathbf{z}). 
% \]
% In this case, the distribution of $\mathbf{z}$ can be assumed to be a pre-defined distribution, e.g. Gaussian $\mathcal{N}(\mathbf{0},\mathbf{I})$, as suggested by \cite{vibg}. 

The bound \eqref{eq:mi2} is obtained by assuming a parametric distribution for $p(\mathbf{z})$.
In this paper, we do not make such an assumption. 
%  I(\mathbf{x};\mathbf{z})\simeq \frac{1}{N} \sum \limits_{i=1}^N \mathbb{E}_{\mathbf{z}\sim p(\mathbf{z}|\mathbf{x}_i)}\big[-\log \frac{1}{N}\sum_{j=1}^N p(\mathbf{z}|\mathbf{x}_j) + \log p(\mathbf{z}|\mathbf{x}_i) \big]
Thus, we propose a non-parametric method to estimate entropy $H(\mathbf{z})$. More precisely, we propose to estimate $p(\mathbf{z})$ non-parametrically by computing
\begin{equation}
\begin{split}
p(\mathbf{z})=\sum \limits_{\mathbf{x}}p(\mathbf{z}|\mathbf{x})p(\mathbf{x}) \simeq\frac{1}{N} \sum \limits_{j=1}^N p(\mathbf{z}|{\mathbf{x}}_j).
 \end{split}
 \label{eq:hz}
\end{equation}
By substituting the above equation into $H(\mathbf{z})$ and applying the Jensen inequality, we obtain
\begin{equation}
\begin{split} 
H(\mathbf{z})&\simeq
\frac{1}{N} \sum \limits_{i=1}^N \mathbb{E}_{\mathbf{z}\sim p(\mathbf{z}|\mathbf{x}_i)}\big[-\log \frac{1}{N} \sum \limits_{j=1}^N p(\mathbf{z}|{\mathbf{x}}_j) \big],\\
&\leq \frac{1}{N} \sum \limits_{i=1}^N \mathbb{E}_{\mathbf{z}\sim p(\mathbf{z}|\mathbf{x}_i)}\big[-\frac{1}{N} \sum \limits_{j=1}^N \log  p(\mathbf{z}|{\mathbf{x}}_j) \big].\\
 \end{split}
 \label{eq:hz}
\end{equation}
By employing the equality $p(\mathbf{z}|\mathbf{x})=\mathcal{N}(\mu(\mathbf{x}),\text{diag}(\sigma^2(\mathbf{x})))$ and \eqref{eq:pzx} in order to generate $K$ samples of $\mathbf{z}$, we have the following bound for $H(\mathbf{z})$,
% H(\mathbf{z})\leq\frac{1}{2KN^2} \sum \limits_{i=1}^N 
% \sum\limits_{k=1}^K \sum \limits_{j=1}^N
% \Big( \big(\mu(\mathbf{x}_j)-\mu(\mathbf{x}_i)-\sigma(\mathbf{x}_i)\odot\epsilon_k \big)^T\text{diag}(\sigma^2(\mathbf{x}_j))^{-1}\\
% \big(\mu(\mathbf{x}_j)-\mu(\mathbf{x}_i)-\sigma(\mathbf{x}_i)\odot\epsilon_k \big)  
%  -\log |2\pi\text{diag}(\sigma^2(\mathbf{x}_j))|    \Big)
\begin{equation}
% \small
\begin{split} 
H(\mathbf{z})\leq
\frac{1}{2KN^2} \sum \limits_{i=1}^N 
\sum\limits_{k=1}^K \sum \limits_{j=1}^N
\Big( \big(\mu(\mathbf{x}_j)-\mu(\mathbf{x}_i)-\sigma(\mathbf{x}_i)\odot\epsilon_k \big)^2 \oslash \sigma^2(\mathbf{x}_j) 
+\\
\log \big|2\pi\text{diag}(\sigma^2(\mathbf{x}_j))\big|    \Big).
\end{split}
\label{eq:lhz}
\end{equation}
where $\epsilon_k\sim \mathcal{N}(\mathbf{0},\mathbf{I})$ and $(\cdot)^2$ is element-wise square, $\oslash$ is element-wise division of vectors\footnote{As a simplification for  $\big(\mu(\mathbf{x}_j)-\mu(\mathbf{x}_i)-\sigma(\mathbf{x}_i)\odot\epsilon_k\big)^T \text{diag}(\sigma^2(\mathbf{x}_j))^{-1} \big(\mu(\mathbf{x}_j)-\mu(\mathbf{x}_i)-\sigma(\mathbf{x}_i)\odot\epsilon_k\big)$}.
Since conditional entropy $H(\mathbf{z}|\mathbf{x})$ can be analytically computed by
\begin{equation}
\begin{split} 
H(\mathbf{z}|\mathbf{x})&\simeq \frac{-1}{N} \sum \limits_{i=1}^N \mathbb{E}_{\mathbf{z}\sim p(\mathbf{z}|\mathbf{x}_i)}\big[\log  p(\mathbf{z}|\mathbf{x}) \big],\\
&=\frac{1}{2N}\sum\limits_{i=1}^N \log |2\pi\text{diag}(\sigma^2(\mathbf{x}_i))|,
 \end{split}
 \label{eq:hzx}
\end{equation}
Combining \eqref{eq:lhz} and \eqref{eq:hzx}, we can compute the upper bound for the mutual information by
\begin{equation}
% \small
\begin{split} 
I(\mathbf{z};\mathbf{x})\leq \frac{1}{2KN^2} \sum \limits_{i=1}^N 
\sum\limits_{k=1}^K \sum \limits_{j=1}^N
\Big( \big(\mu(\mathbf{x}_j)-\mu(\mathbf{x}_i)-\sigma(\mathbf{x}_i)\odot\epsilon_k \big)^2 \oslash \sigma^2(\mathbf{x}_j) \Big). 
 \end{split}
 \label{eq:lmi2}
\end{equation}

Finally, by using \eqref{eq:lmi2} in our optimization objective in \eqref{eq:rd2}, we have the following objective for training an AE,
\begin{multline}
% \begin{split} 
\min_{\mu,\sigma,g}\quad \frac{1}{2KN^2} \sum \limits_{i=1}^N 
\sum\limits_{k=1}^K \sum \limits_{j=1}^N
\Big( d(\mathbf{x}_i,g(\mu(\mathbf{x}_i)+\sigma(\mathbf{x}_i)\odot\epsilon_k)) 
 \,+ \\ \, \beta\, 
\big(\mu(\mathbf{x}_j)-\mu(\mathbf{x}_i)-\sigma(\mathbf{x}_i)\odot\epsilon_k \big)^2 \oslash \sigma^2(\mathbf{x}_j) 
  \Big) .
%  \end{split}
 \label{eq:obj}
\end{multline}
As we can see from the formula, minimizing the mutual information term can be considered as reducing the pair-wise distances between samples in the encoding space. The scheme of reducing pair-wise distances between samples are depicted as \textit{information potentials} in \cite{eminon}. By reducing the pair-wise distances between samples in the encoding space, the information potential aims to map two different samples from the input space to a same sample in the encoding space. If all samples are mapped to a single point in the encoding space, $I(\mathbf{x};\mathbf{z})$ will be minimized. However, the reconstruction error will be maximized. On the other hand, solely reducing the reconstruction error term in \eqref{eq:obj} will map samples to different points in the encoding space, which may result in a trivial solution of identity mapping. Hence, mutual information can serve as a regularization for AEs.
These two errors are trade-offed by the coefficient $\beta$.

Although the motivation of VAEs is not based on information theory, the regularization term employed by VAEs has the same form as \eqref{eq:mi2}, which is a parametric estimation of $H(\mathbf{z})$. As we can see from \eqref{eq:mi2}, minimizing mutual information in \eqref{eq:mi2} can be considered as minimizing the distances in the encoding space between all samples to a center point which is $\mathbf{0}$. 
Thus, from an information theoretic perspective, the difference between VAEs and IPAEs is that, the regularization of IPAEs aims to map any \textit{two} different samples from input space to a same point in the encoding space while the regularization of VAEs is to map \textit{all} samples from input space to a same point in the encoding space.

\section{Related Work}

\paragraph{AEs with Stochastic Encoding:}
Our proposed objective function is identical to that of Variational Auto-Encoders (VAEs) \cite{vae}, if $\beta=1$ and $p(\mathbf{z})$ is assumed to be a pre-defined distribution. However, the motivation for constructing the corresponding objective function is quite different.
VAEs are generative models.
A common approach used to train VAEs is to maximize the marginal log-likelihood of observed samples. $p(\mathbf{z})$ utilized in VAEs is a prior distribution that generates $\mathbf{x}$. If the real prior distribution is far way from the assumed prior distribution, then VAEs cannot learn \textit{useful} representations.
% The assumption here is each sample $\mathbf{x}$ is generated by random process $p(\mathbf{x}|\mathbf{z})p(\mathbf{z})$.
Denoising Auto-Encoders (DAEs) \cite{dae} avoid the trivial solutions by extracting encoding representations from a noise-corrupted input, and aim to reconstruct clean inputs from the extracted encoding representations. With the denoising criteria, DAEs can learn useful edge detectors from images. In our model, noise is added to the encode variable as shown in \eqref{eq:pzx}. 

\paragraph{Mutual Information in Representation Learning:}
Previous works on non-parametric estimation of mutual information reduce the aforementioned problem to estimation of Renyi Entropy, e.g. \cite{eminon,eminon2,rateae}. In \cite{eminon}, a
Parzen window model was employed for density estimation. In \cite{eminon2,rateae}, the entropy estimation is based on kernel methods that can avoid explicit estimation of probability density. 
Although non-parametric methods do not make assumptions for the distributions, the disadvantages are also obvious.
Non-parametric methods usually have to explore all pairs of samples. Parametric methods can be computationally less complex by making assumptions to probability density functions. For instance, variables were assumed to be drawn from Gaussian distributions in \cite{gib}. In addition, analytic solutions of the information bottleneck problem \cite{ib} can be found considering affine transformations of variables. \cite{vibg} adopted a stochastic encoding method using a variational method for computation of the bound for the mutual informatio when variables are transformed in multi-layer deep networks endowed with non-linearity activation functions. Our proposed method of estimating mutual information can be considered as a hybrid of parametric (for $H(\mathbf{z}|\mathbf{x})$) and non-parametric (for $H(\mathbf{z})$) models.

\paragraph{Other Regularization Methods for AEs:} A well known regularization method used for training AEs is mapping a representation space of encoding variables to a lower dimensional space than that of an input space. Then, the lower-dimensional encoding variable can be considered as an under-complete representation of the input. 
% It is expected to produce non-zero reconstruction errors and force the Auto-Encoder to avoid the trivial solutions.
Unlike the methods that perform regularization using lower dimensionality, sparse coding \cite{sparsecoding} employs a higher dimensionality on the encoding space to extract over-complete representations. Using higher-dimensional encoding variables, sparse coding methods impose sparsity constraints to AEs in order to avoid the trivial solutions. Note that, sparsity constraints cause inhibition of a larger number of hidden units on the output layer of the encoder network. Thus, AEs trained using sparsity constraints produce many zero activation values. As a result, the sparse coding can be considered as an implicit compression method used to avoid trivial solutions. In our proposed method, there is no dimensionality constraint. 
Winner-Take-All AEs \cite{wta} can also be considered as a sparsity regularization method. The sparsity is obtained by keeping the hidden units with top activation values and inhibiting the rest of units. Winner-Take-All scheme can learn Gabor-like filters in convolutional or fully connected AEs.

\section{Experiments}

In this section, we empirically examine the proposed Information Potential Auto-encoder (IPAE) method in comparison with VAEs. For a fair comparison of the results, the coefficient $\beta$ is also used in VAEs to trade-off regularization and reconstruction, resulting in a similar model suggested in $\beta$-VAEs \cite{betavae}. While generating $\mathbf{z}$ samples, $K$ is set as $1$ for both AEs in all experiments.

\subsection{Toy Dataset}

We first conduct experimental analyses on a toy dataset. We generated the toy dataset using a mixture of $25$ Gaussian distributions in a two-dimensional space. Each Gaussian has the same covariance matrix $[ \begin{smallmatrix} 0.1&0\\ 0&0.1 \end{smallmatrix}]$, but with different means. We generate $200$ samples from each Gaussian and obtain $5000$ samples.
The target is to reveal the underlying structure of the data, i.e. to estimate $25$ centers of Gaussian distributions. The best result is obtained when all samples drawn from the same Gaussian collapse to the same point.
If we consider $25$ Gaussian distributions as distributions of $25$ classes, the best result provides a best information bottleneck trade-off \cite{ib} by keeping useful information (class membership) and removing noise.

\begin{figure}[t]
\captionsetup[subfigure]{justification=centering}
% \centering
\begin{subfigure}[b]{.33\textwidth}
\centering
  \includegraphics[height=38mm,width=.95\linewidth]{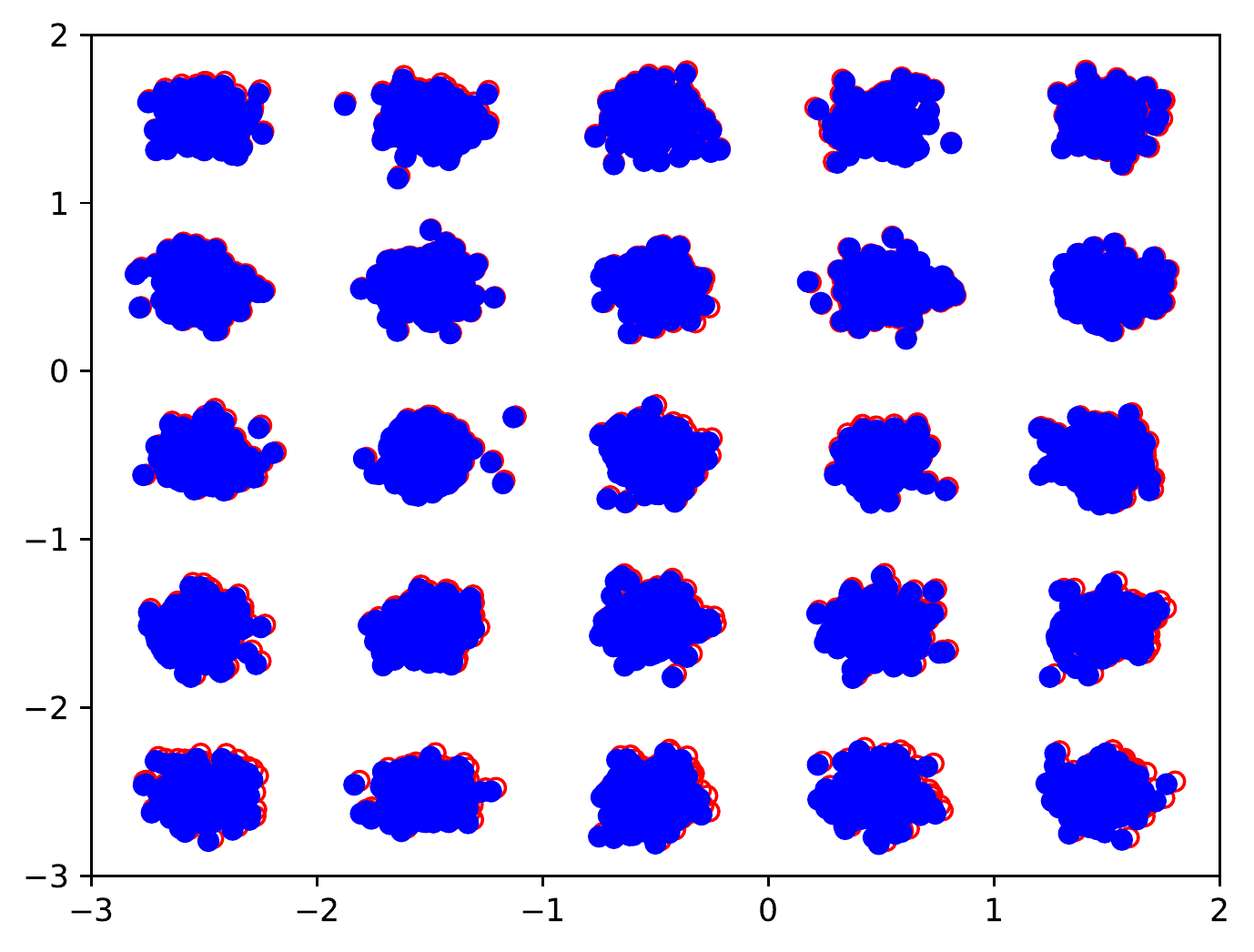}
  \caption{VAE, $\beta=0.0001$\\$\mathcal{E}=0.00972$}
  \label{fig:1a}
\end{subfigure}%
\begin{subfigure}[b]{.33\textwidth}
\centering
  \includegraphics[height=38mm,width=.95\linewidth]{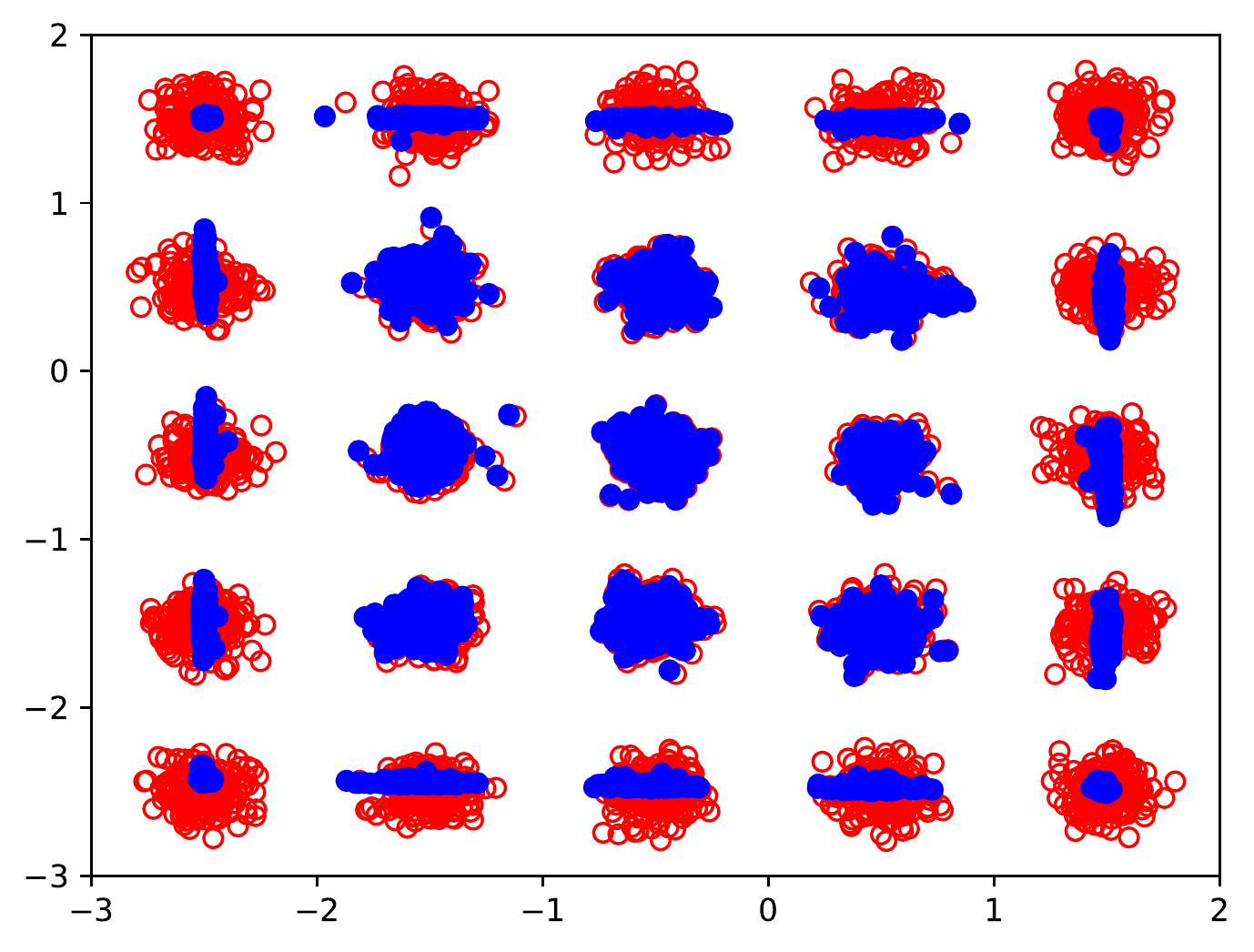}
  \caption{VAE, $\beta=0.1$\\$\mathcal{E}=0.00728$}
  \label{fig:1b}
\end{subfigure}%
\begin{subfigure}[b]{.33\textwidth}
\centering
  \includegraphics[height=38mm,width=.95\linewidth]{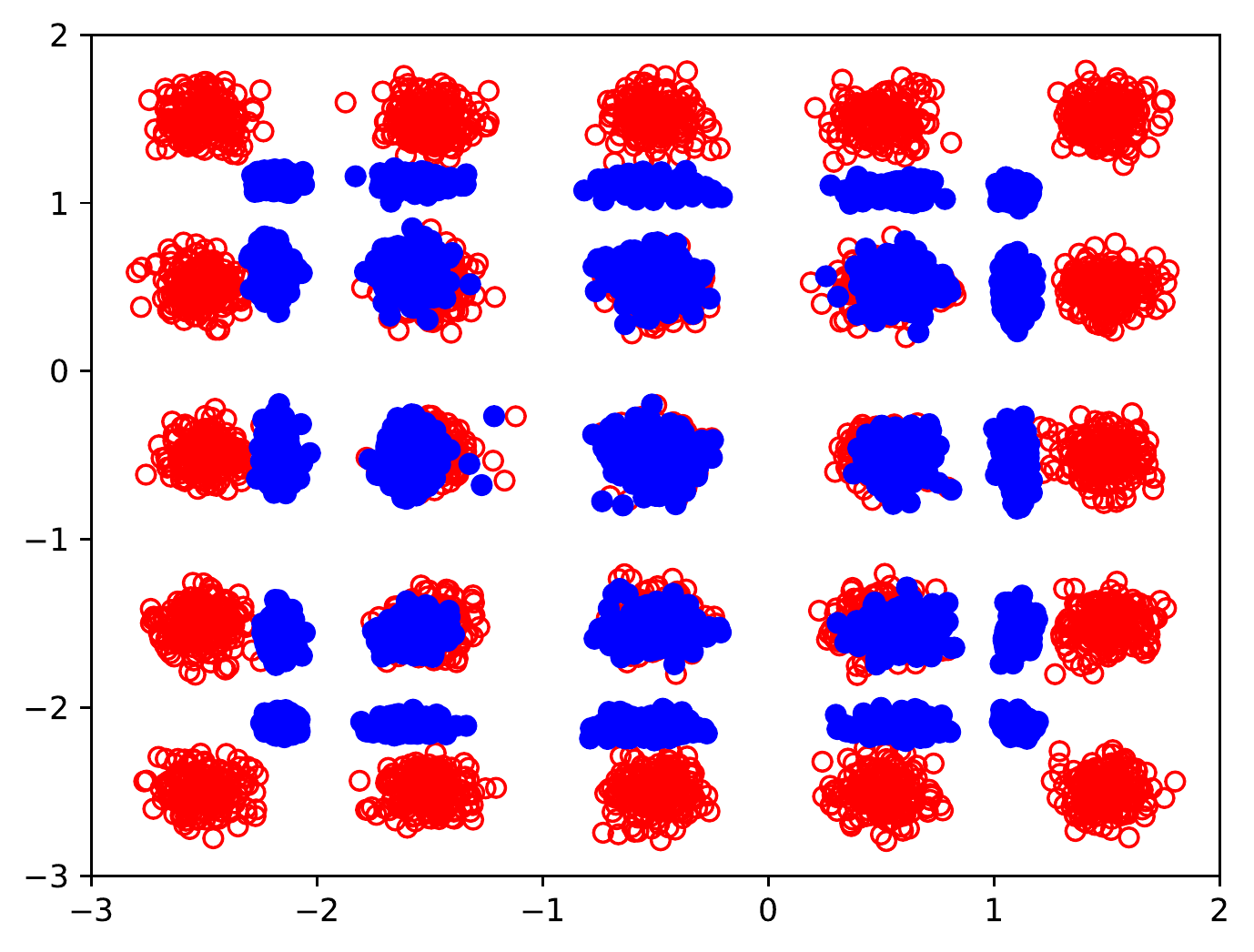}
  \caption{VAE, $\beta=0.5$\\$\mathcal{E}=0.0653$}
  \label{fig:1c}
\end{subfigure}%
\\

\begin{subfigure}[b]{.33\textwidth}
\centering
  \includegraphics[height=38mm,width=.95\linewidth]{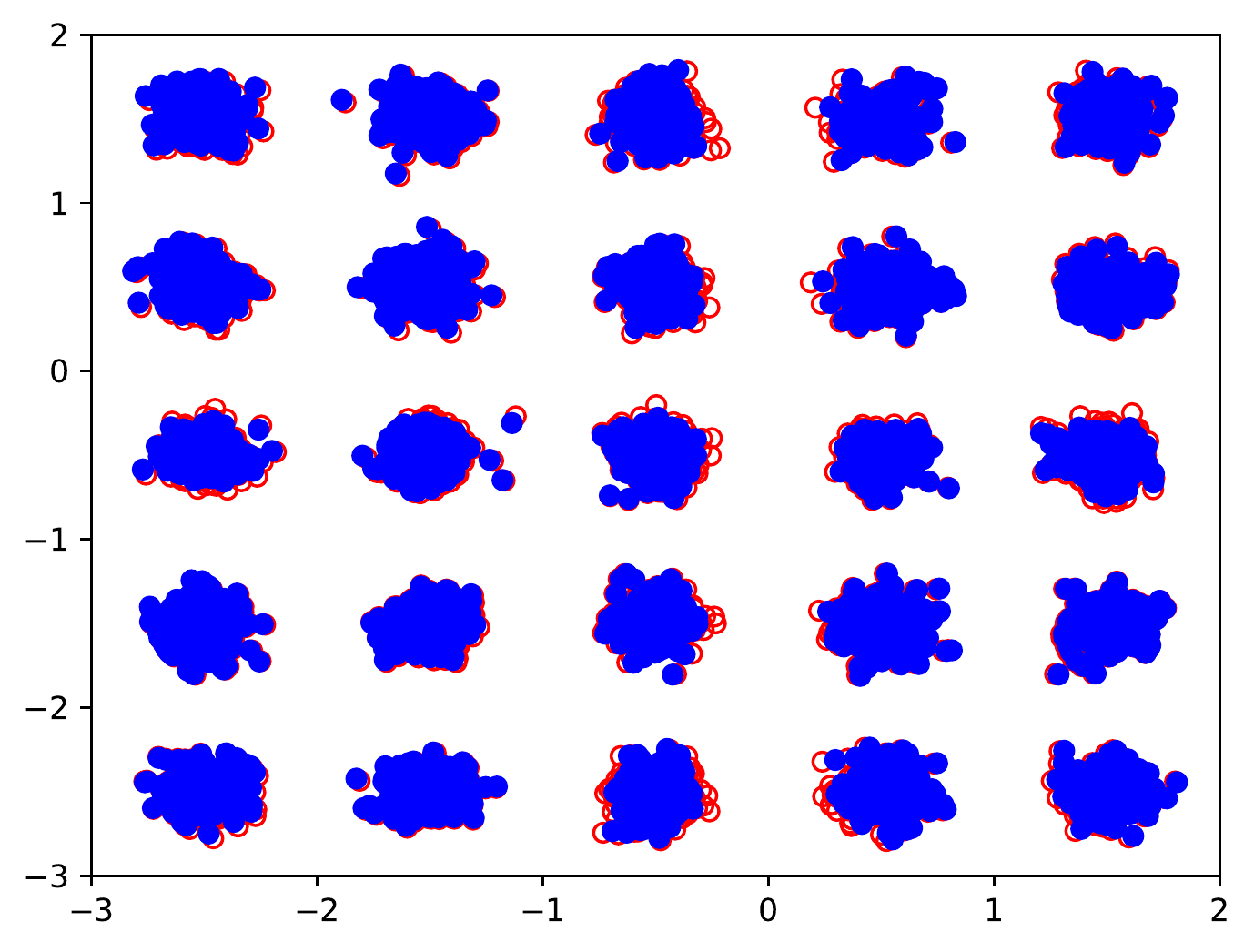}
  \caption{IPAE, $\beta=0.00001$\\$\mathcal{E}=0.00899$}
  \label{fig:1d}
\end{subfigure}%
\begin{subfigure}[b]{.33\textwidth}
\centering
  \includegraphics[height=38mm,width=.95\linewidth]{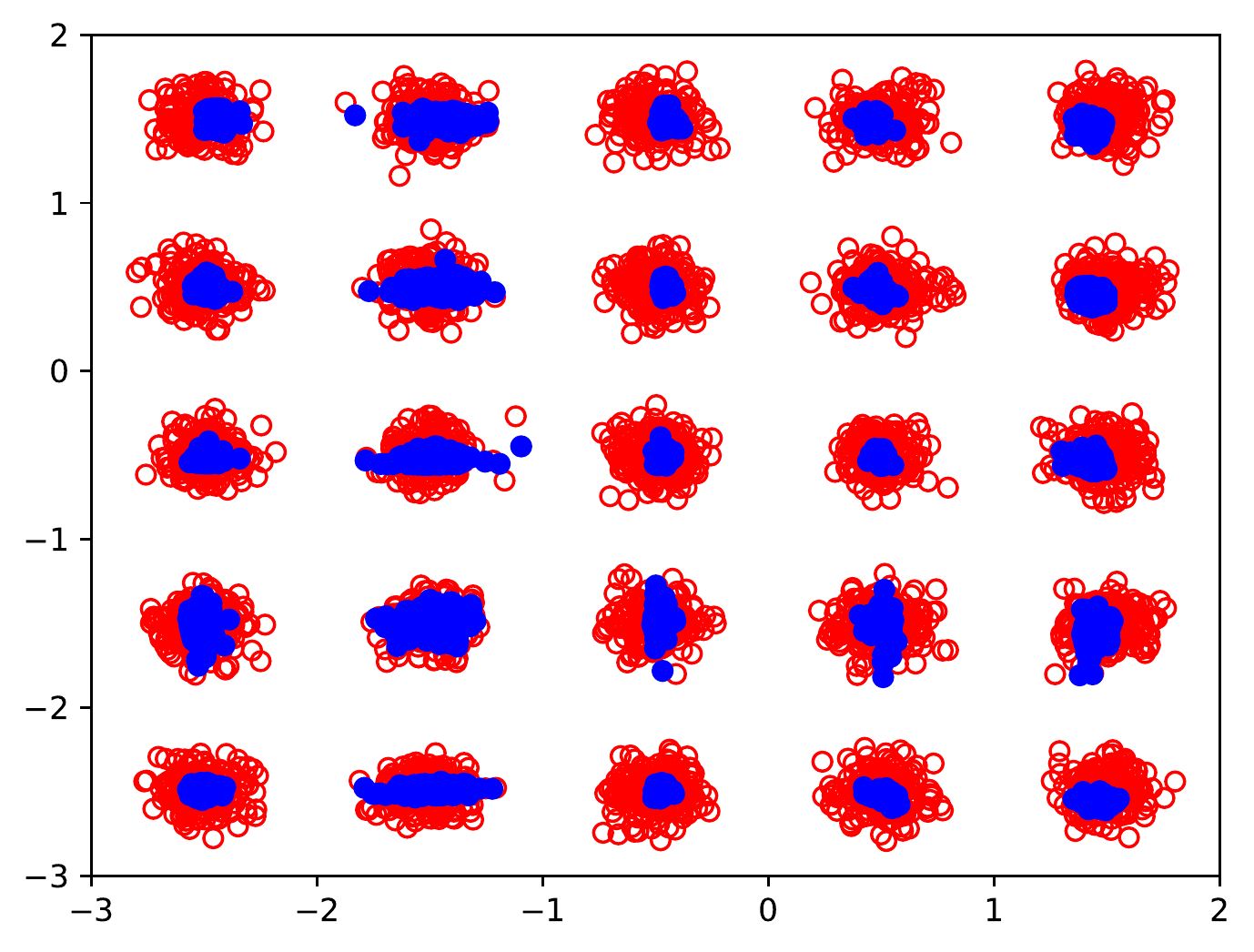}
  \caption{IPAE, $\beta=0.0005$\\$\mathcal{E}=0.00234$}
  \label{fig:1e}
\end{subfigure}%
\begin{subfigure}[b]{.33\textwidth}
\centering
  \includegraphics[height=38mm,width=.95\linewidth]{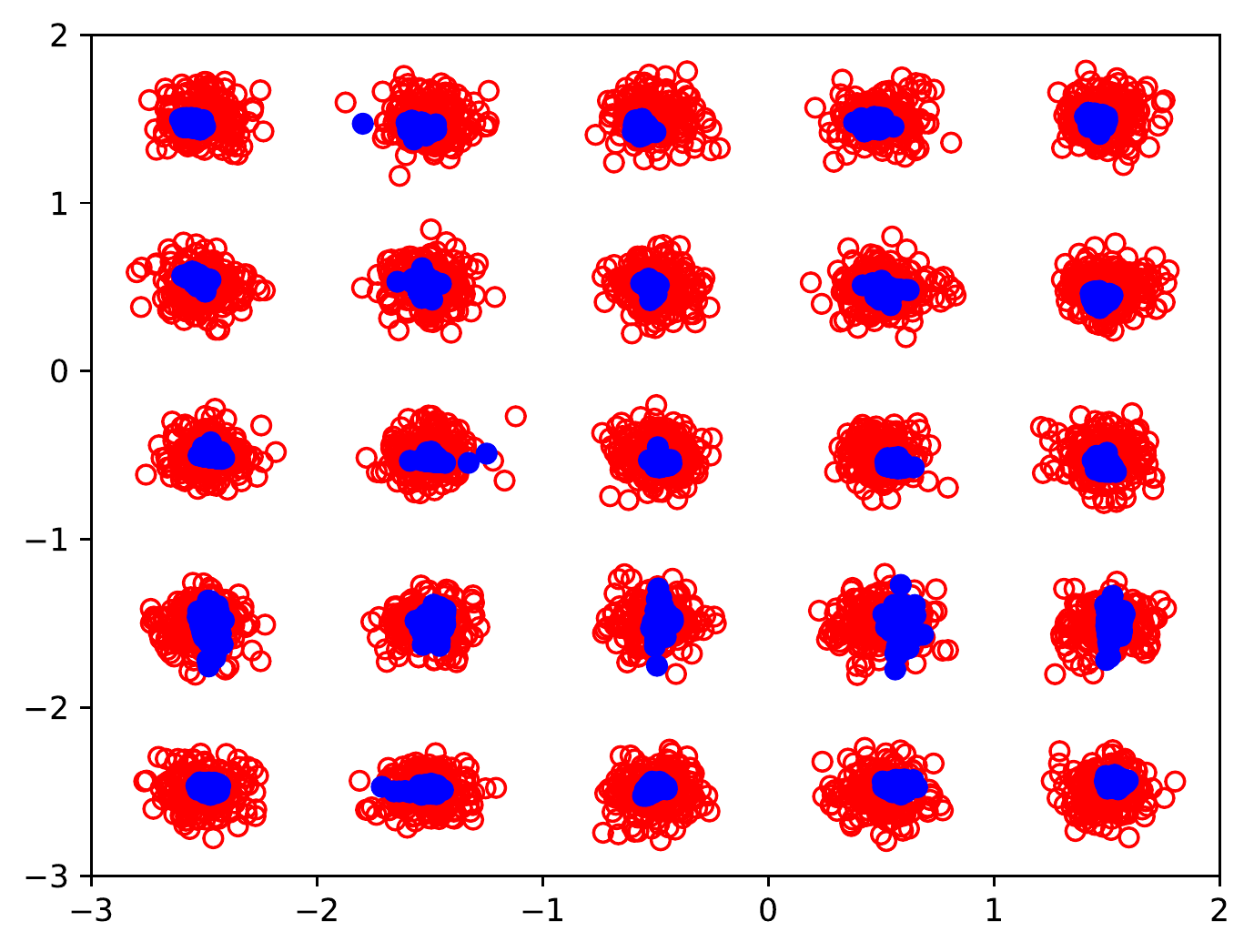}
  \caption{IPAE, $\beta=0.001$\\$\mathcal{E}=0.00204$}
  \label{fig:1f}
\end{subfigure}%

\caption{Reconstruction results obtained using VAE and IPAE for different $\beta$. In addition to the values illustrated in the figure, other values of $\beta$ are also used to train VAEs, but cannot obtain better results than that of the IPAE for $\beta=0.001$.
The red circles denote input data and the blue dots denote reconstructed data.}
\label{fig:toy}
\end{figure}

% consists of $15000$ samples in a $3$-dimensional space. Each sample $\mathbf{x}$ is generated by the following two steps of a random process:
% \begin{enumerate}
%     \item Generate a value of $h$ from uniform distribution in $[0,180]$.
%     \item The value of each dimension of $\mathbf{x}$ is generated by $\mathbf{x}[1]=cos(h)+\epsilon$, $\mathbf{x}[2]=sin(h)+\epsilon$, and $\mathbf{x}[3]=sin(h)+\epsilon$, where $\mathbf{x}[i]$ denotes the  $i^{th}$ dimension, and $\epsilon$ is a Gaussian noise $\mathcal{N}(0,0.05)$. 
%     % a noise $\epsilon_2$ from a  distribution  
%     % \begin{equation*}
%     % \begin{split}
%     % \mathbf{x}[1]=cos(\epsilon_1)+\epsilon_2\\
%     % \mathbf{x}[2]=sin(\epsilon_1)+\epsilon_2\\
%     % \mathbf{x}[3]=sin(\epsilon_1)+\epsilon_2\\
%     %  \end{split}
%     %  \label{eq:rp}
%     % \end{equation*}
% \end{enumerate}
% The dataset is further normalized such that the range of values in each dimension is in $[0,1]$. The distribution of the given dataset in the input space is visualized in Figure \ref{fig:toyinput}. 
 
 The network architecture, which was used to construct encoder and decoder, is as follows:
    \begin{equation*}
    \begin{split}
    \mathbf{x}\xrightarrow{\text{relu}} 2048 \rightarrow
    \begin{cases}
        \mu(\mathbf{x})\\
        \sigma(\mathbf{x})\\
    \end{cases}
\dasharrow
\mathbf{z}\xrightarrow{\text{relu}} 2048
\rightarrow 2
     \end{split}
     \label{eq:toyipae}
    \end{equation*}
where "$\rightarrow$" denotes a fully connected layer, on top of which an activation function was used. The number given after each right arrow denotes the number of output units of the fully connected layer. $\mu(\mathbf{x})$ and $\sigma(\mathbf{x})$ are both $16$-dimensional. 
"$\dasharrow$" denotes a stochastic sampling used according to $\mathcal{N}(\mu(\mathbf{x}),\text{diag}(\sigma^2(\mathbf{x})))$.
We use mean square error for a distortion function $d(\cdot,\cdot)$ in our objective \eqref{eq:obj} for this task. 

% cross entropy loss to measure reconstruction error,
% \begin{equation}
% \begin{split}
% d(\mathbf{x},\tilde{\mathbf{x}})=-sum( \mathbf{x}\log \tilde{\mathbf{x}} + (1-\mathbf{x}) \log (1-\tilde{\mathbf{x}}) )
% \end{split}
% \label{eq:dis}
% \end{equation}   
% where $\tilde{\mathbf{x}}$ is the reconstructed vector and $sum(\cdot)$ computes the summation over all dimensions. \eqref{eq:dis} can also be used in VAE, in which $p(\mathbf{x}|\mathbf{z})$ is considered as following a Bernoulli distribution.
 
For all trials of this dataset, we use the following training procedure. Adam \cite{kingma2014adam} optimization method is employed for training. The initial learning rate is set as $0.001$. Batch size is $512$. The training is finished after $5000$ batches.

A comparative result obtained for IPAE and VAE is shown in Figure \ref{fig:toy}. In this experiment, the maximal value for $j$ used in the objective \eqref{eq:obj} is set as $1$ instead of $N$.
To quantify the \textit{performance} of reconstruction results, we also compute the average euclidean distance (denoted as $\mathcal{E}$ in Figure 1) of all reconstructed samples to the means of Gaussian from which they were drawn. The smaller $\mathcal{E}$ is better.
As we can see from the figure, with small $\beta$ value, both AEs failed to find the underlying structure by simply learning an identity mapping between input output (see \ref{fig:1a} and \ref{fig:1d}). As a result, $\mathcal{E}$ value is also large.
As $\beta$ increases, both AEs tend to compress different data points from the input space to the same point in the encoding space, and $\mathcal{E}$ value gets smaller, as shown in \ref{fig:1b} and \ref{fig:1e}. 
Further increasing the $\beta$ value will give a stronger regularization. In the output results (\ref{fig:1c}) of VAEs,
the data points start to deviate from the means, and $\mathcal{E}$ also increases.
On the other hand, IPAE can keep the data points around Gaussian means while compressing them (see Figure 1(f)), and thereby IPAE can obtain a minimal $\mathcal{E}$ value.
The different results obtained by IPAE and VAEs can be interpreted by the different mutual information regularization methods used by two methods. As shown in Section 2, during the learning phase, VAE aims to map all samples to a same point in the encoding space due to employment of parametric mutual information regularization. On the other hand, IPAE aims to map any two samples in the input space to a same point in the encoding space, which gives more degree of freedom to learn the distribution of encoding features.  
Thus, for data drawn from a complex  distribution, such as the proposed $25$ mixture Gaussian distribution, IPAE learns a better data structure compared to VAEs.

% for the same $\beta$ value, IPAE shows stronger compression rate, i.e. smaller $\mathcal{E}$ value. For instance, with $\beta=0.0001$, IPAE can compress the data to smaller blobs as shown in Figure \ref{fig:1e} while the data points reconstructed by VAE show a larger covariance as observed in Figure \ref{fig:1b}.
%  If the value of $\beta$ is too large, the data points start to deviate from the means (see Figure 1(d) and 1(h)) and $\mathcal{E}$ also increases.
% For an appropriate $\beta$ value,  On the other hand, in VAE, some Gaussian blobs can be compressed while some cannot, as we can observe in Figure 1(g). 

The main additional computational cost of IPAE compared to VAEs comes from maximal values of $j$ (notated as $N_j$ in the following) used in objective \eqref{eq:obj}. By setting $N_j=N$, computational cost can be increased. However, our comparative experiments show that $N_j=1$ works just fine. $N_j$ is the number of samples that are used to approximate $p(\mathbf{z})$ in our method. Note that in the inference phase, our model is as fast as VAEs, since $N_j$ is no longer required.

% Hence, we conduct several experiments to show the effect of $N_j$ in order to train IPAEs. The results are shown in Figure 2. 
% As we can see, as $N_j$ increases, the results get better with cost of increasing learning time.

% the distribution of samples in the encoded space and the distribution of reconstructed data are shown in Figure \ref{fig:toy}. As we can see from the figures

\subsection{Experimental Analyses on the MNIST Dataset}

\begin{figure}[t]
\captionsetup[subfigure]{justification=centering}
% \centering
\begin{subfigure}[b]{.33\textwidth}
\centering
  \includegraphics[height=38mm,width=.9\linewidth]{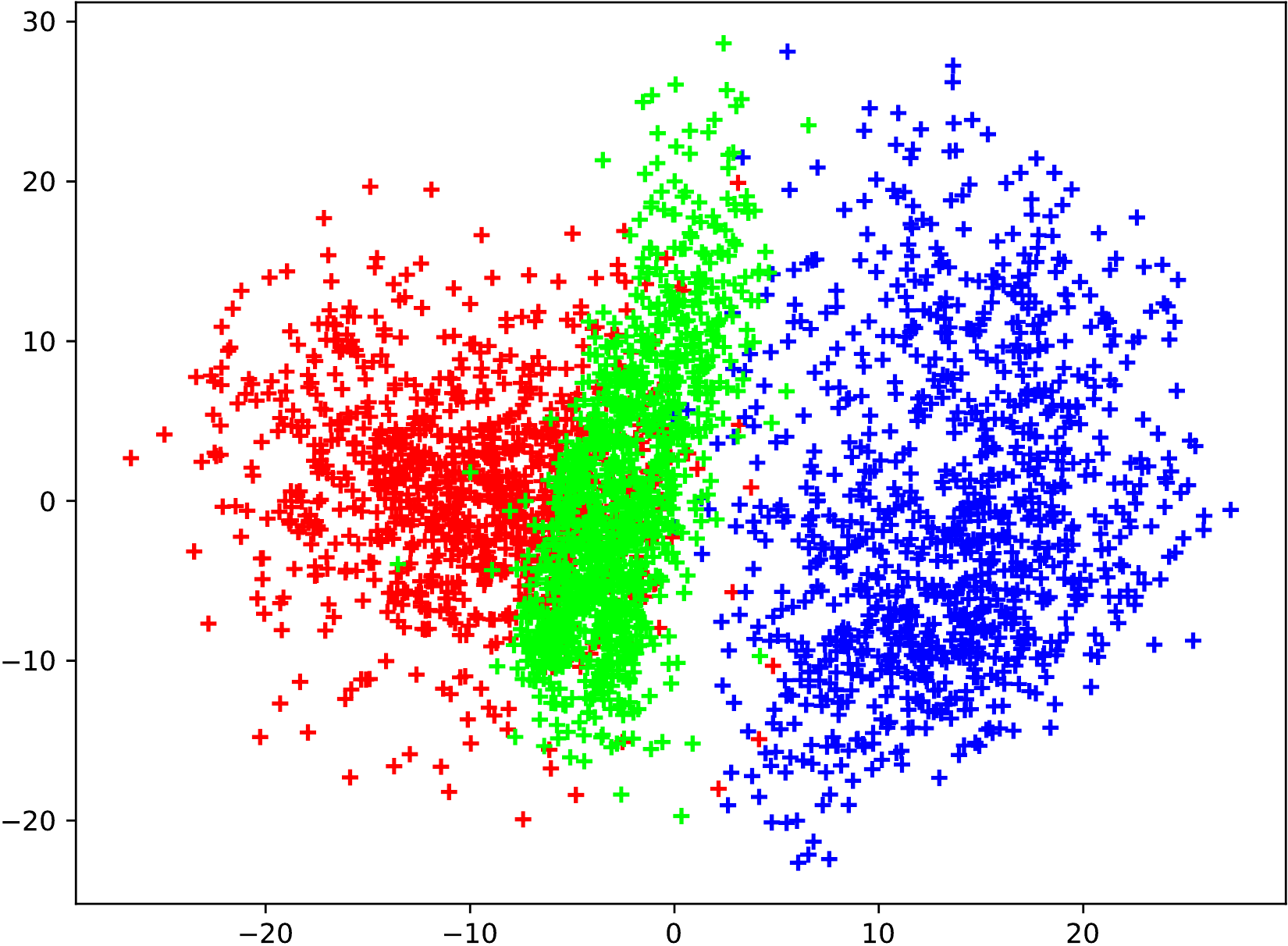}
  \caption{VAE, $\beta=0.001$\\ $err=0.82\pm0.14$.}
  \label{fig:2a}
\end{subfigure}%
\begin{subfigure}[b]{.33\textwidth}
\centering
  \includegraphics[height=38mm,width=.9\linewidth]{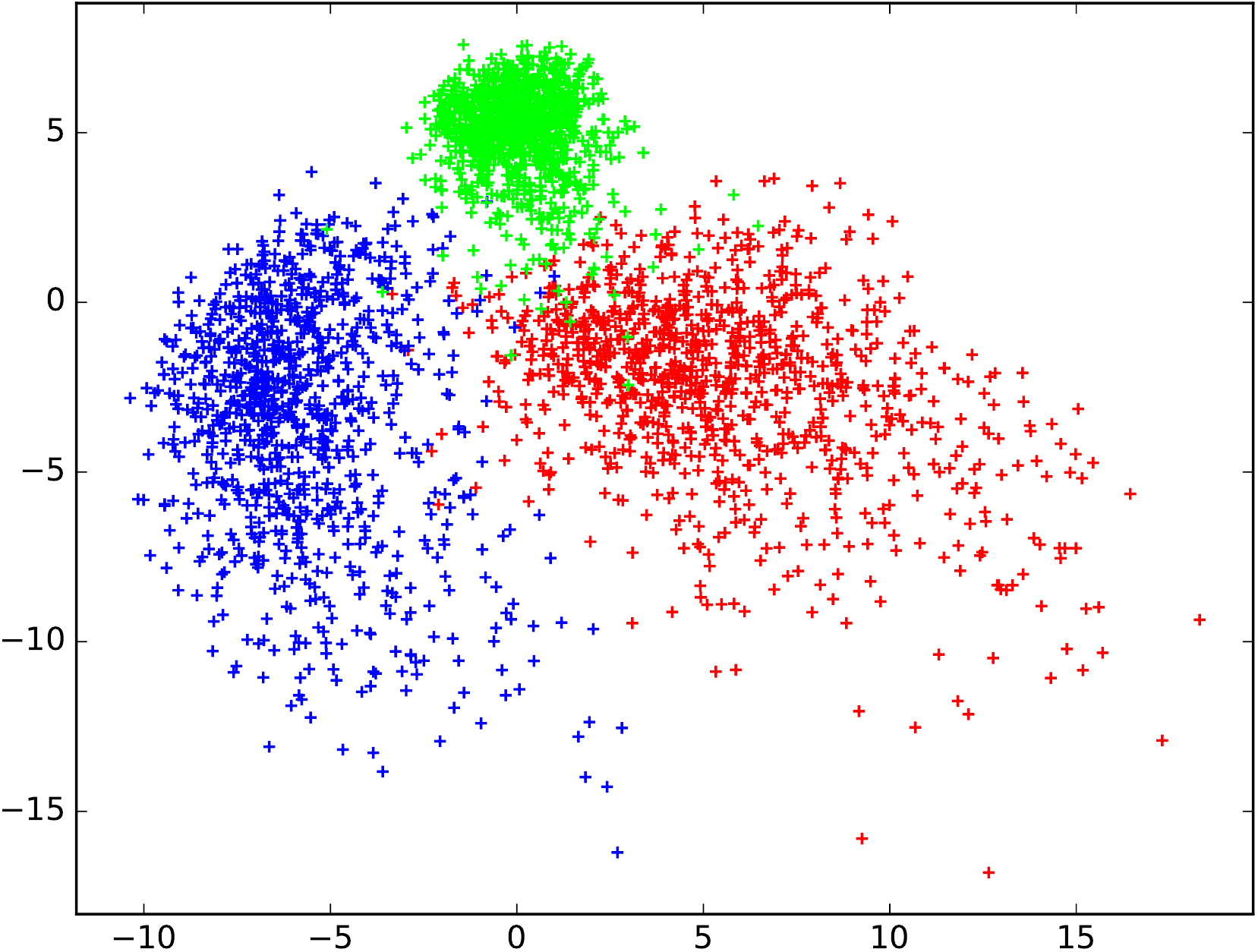}
  \caption{VAE, $\beta=0.01$\\ $err=0.93\pm0.17$.}
  \label{fig:2b}
\end{subfigure}%
\begin{subfigure}[b]{.33\textwidth}
\centering
  \includegraphics[height=38mm,width=.9\linewidth]{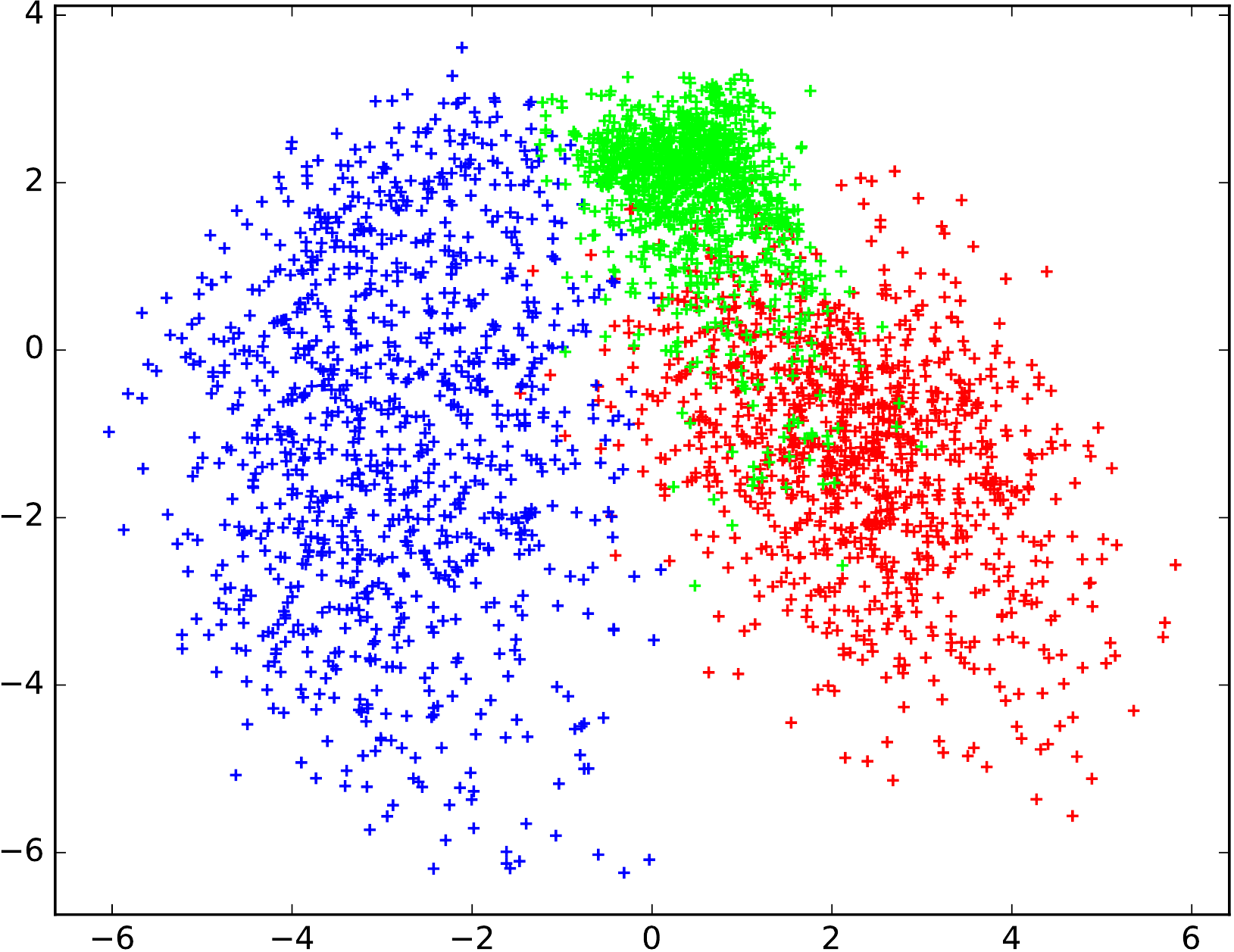}
  \caption{VAE, $\beta=0.1$\\ $err=0.82\pm 0.08$.}
  \label{fig:2c}
\end{subfigure}%
\\
\begin{subfigure}[b]{.33\textwidth}
\centering
  \includegraphics[height=38mm,width=.9\linewidth]{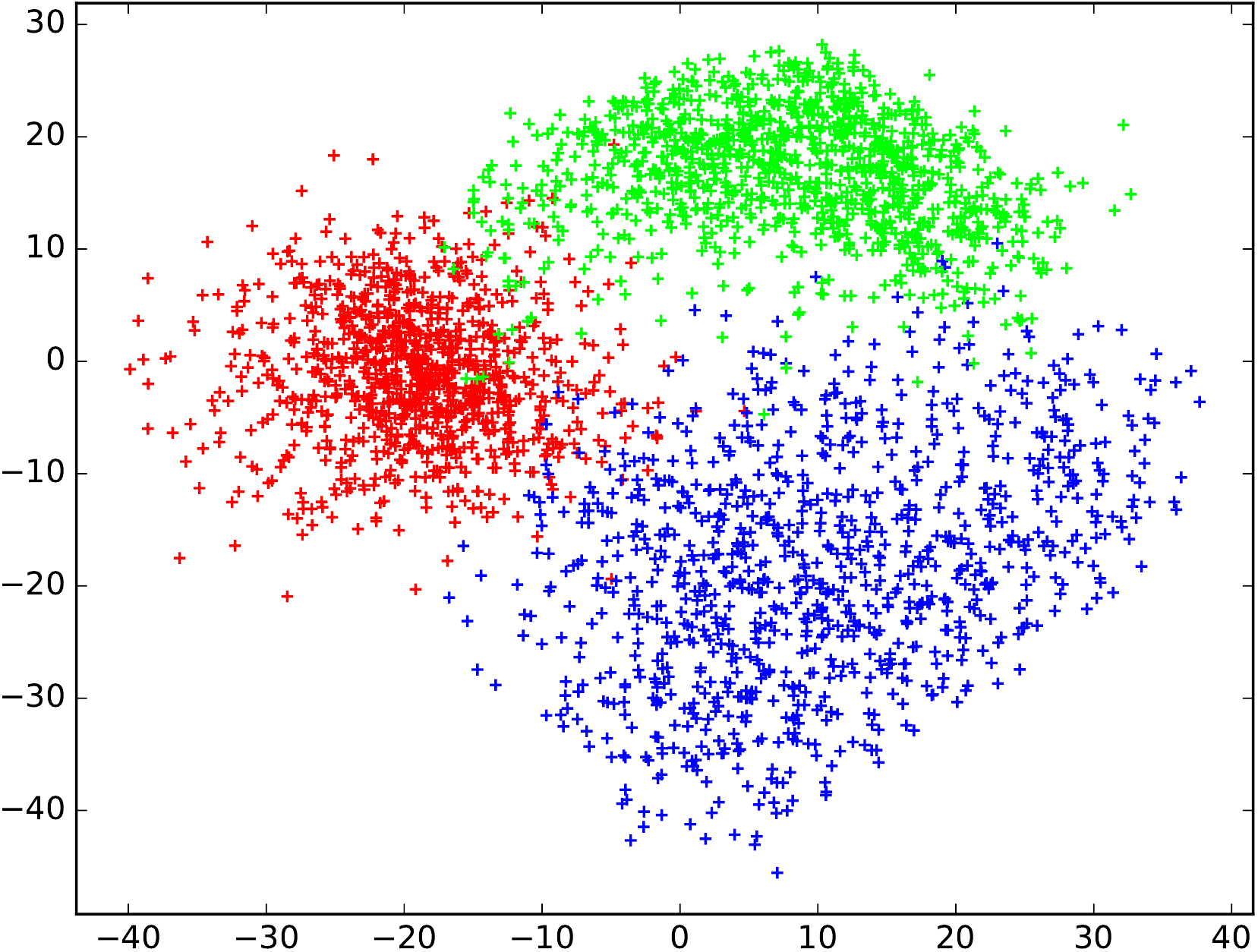}
  \caption{IPAE, $\beta=0.00001$\\ $err=\mathbf{0.73\pm0.22}$.}
  \label{fig:2d}
\end{subfigure}%
\begin{subfigure}[b]{.33\textwidth}
\centering
  \includegraphics[height=38mm,width=.9\linewidth]{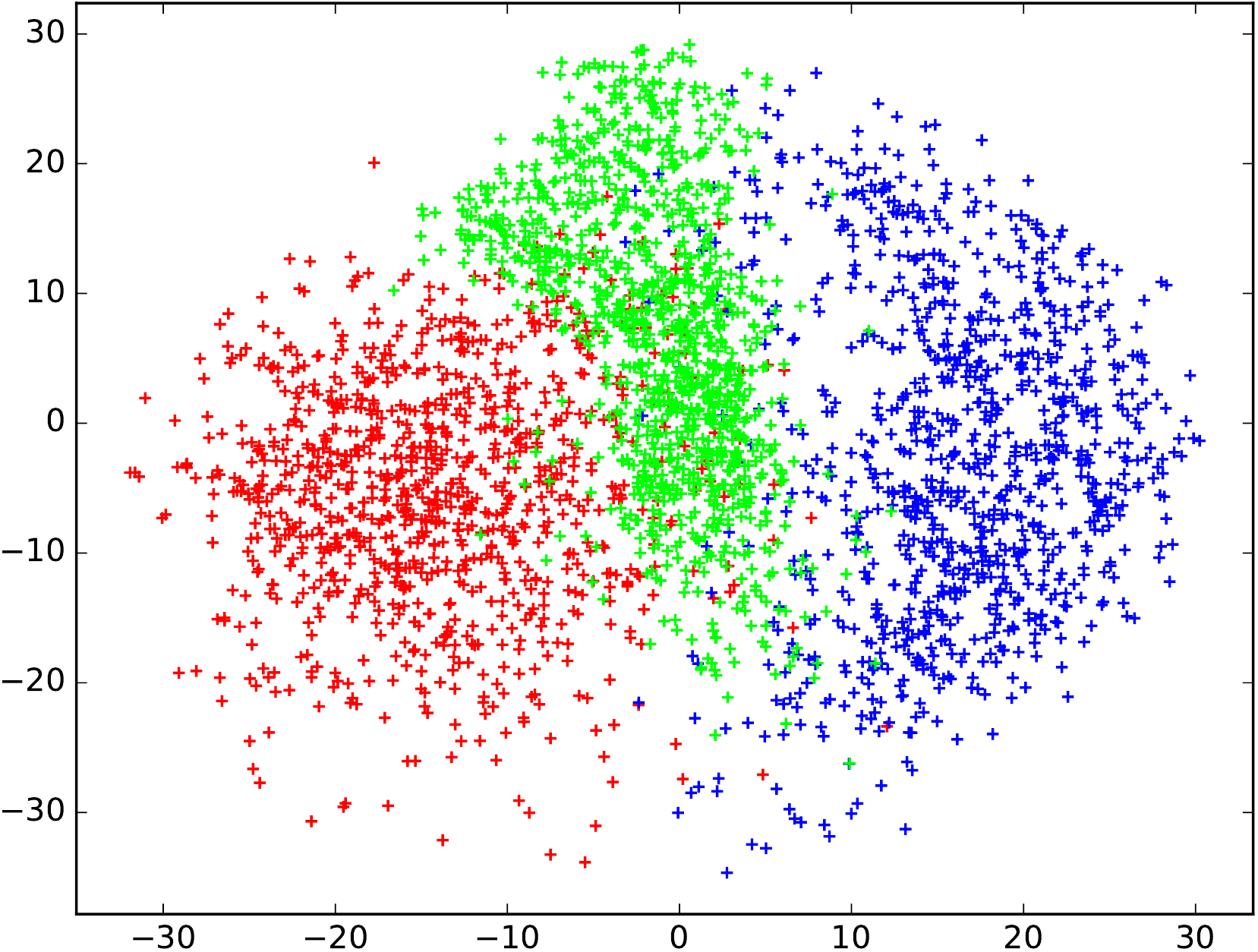}
  \caption{IPAE, $\beta=0.01$\\ $err=1.58\pm0.50$.}
  \label{fig:2e}
\end{subfigure}%
\begin{subfigure}[b]{.33\textwidth}
\centering
  \includegraphics[height=38mm,width=.9\linewidth]{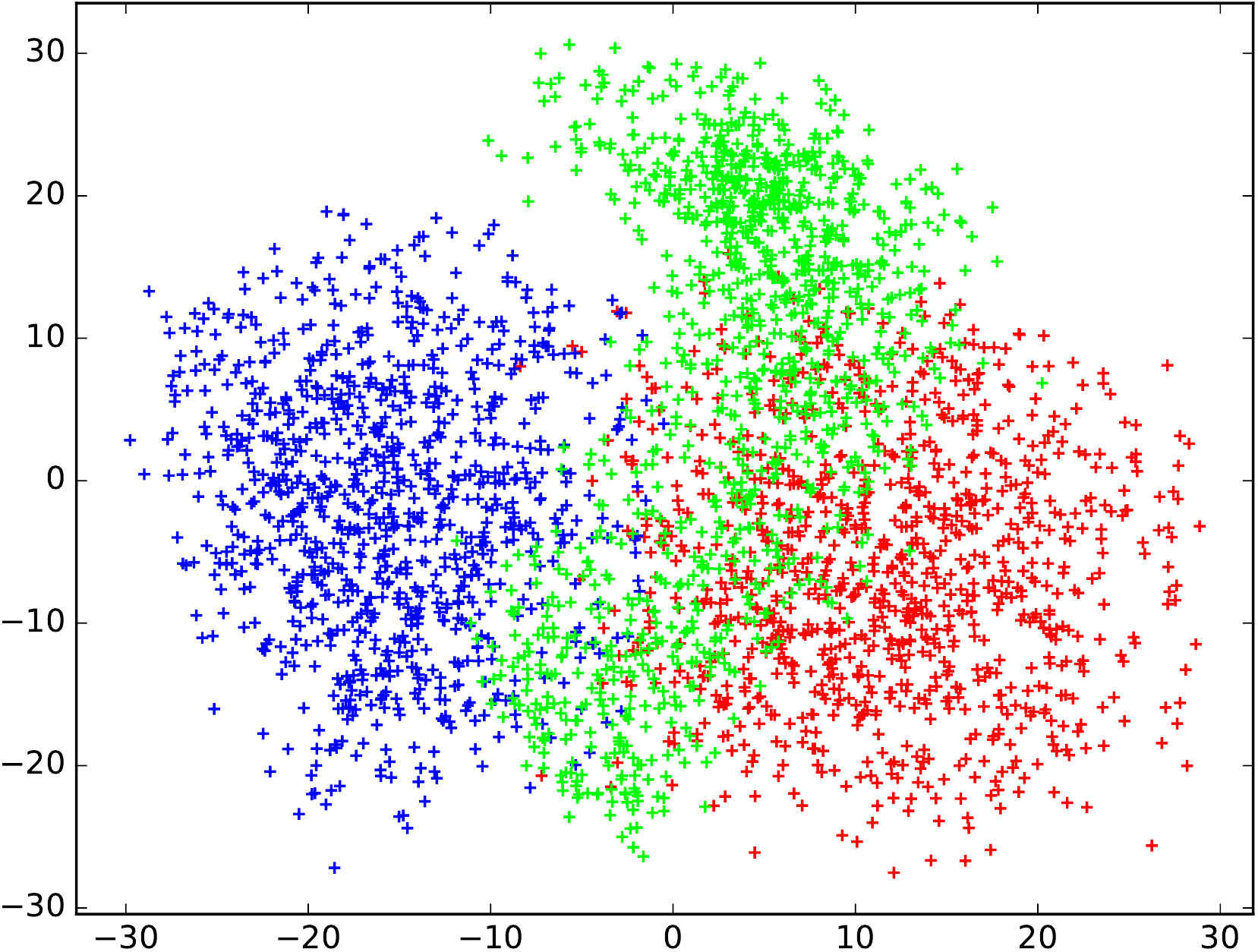}
  \caption{IPAE, $\beta=0.1$\\ $err=1.80\pm0.40$.}
  \label{fig:2f}
\end{subfigure}%
\\
\begin{subfigure}[b]{.33\textwidth}
\centering
  \includegraphics[height=38mm,width=.9\linewidth]{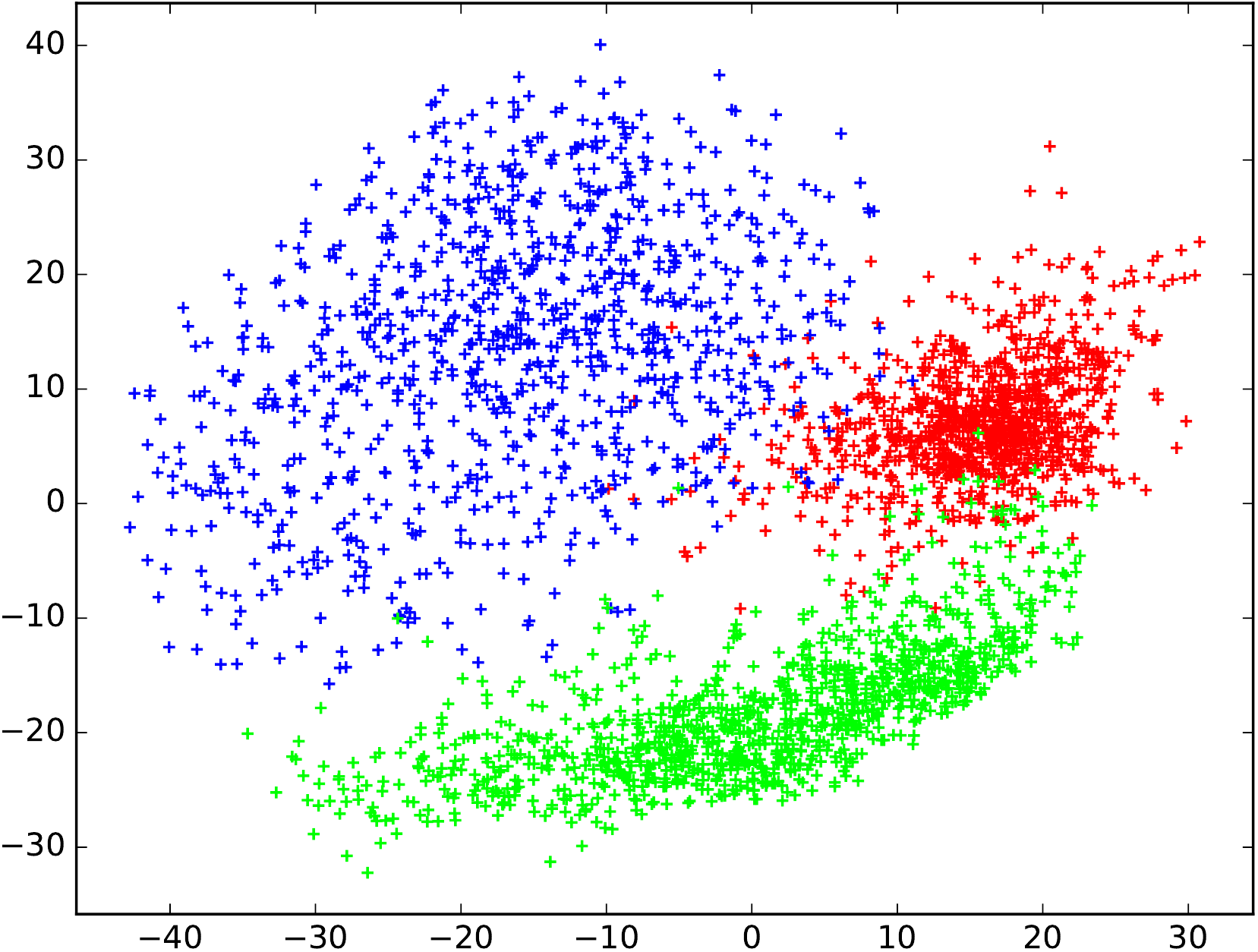}
  \caption{IPAE, $\beta=0.00001$, $N_j=8$\\ $err=\mathbf{0.70\pm0.15}$.}
  \label{fig:2g}
\end{subfigure}%
\begin{subfigure}[b]{.33\textwidth}
\centering
  \includegraphics[height=38mm,width=.9\linewidth]{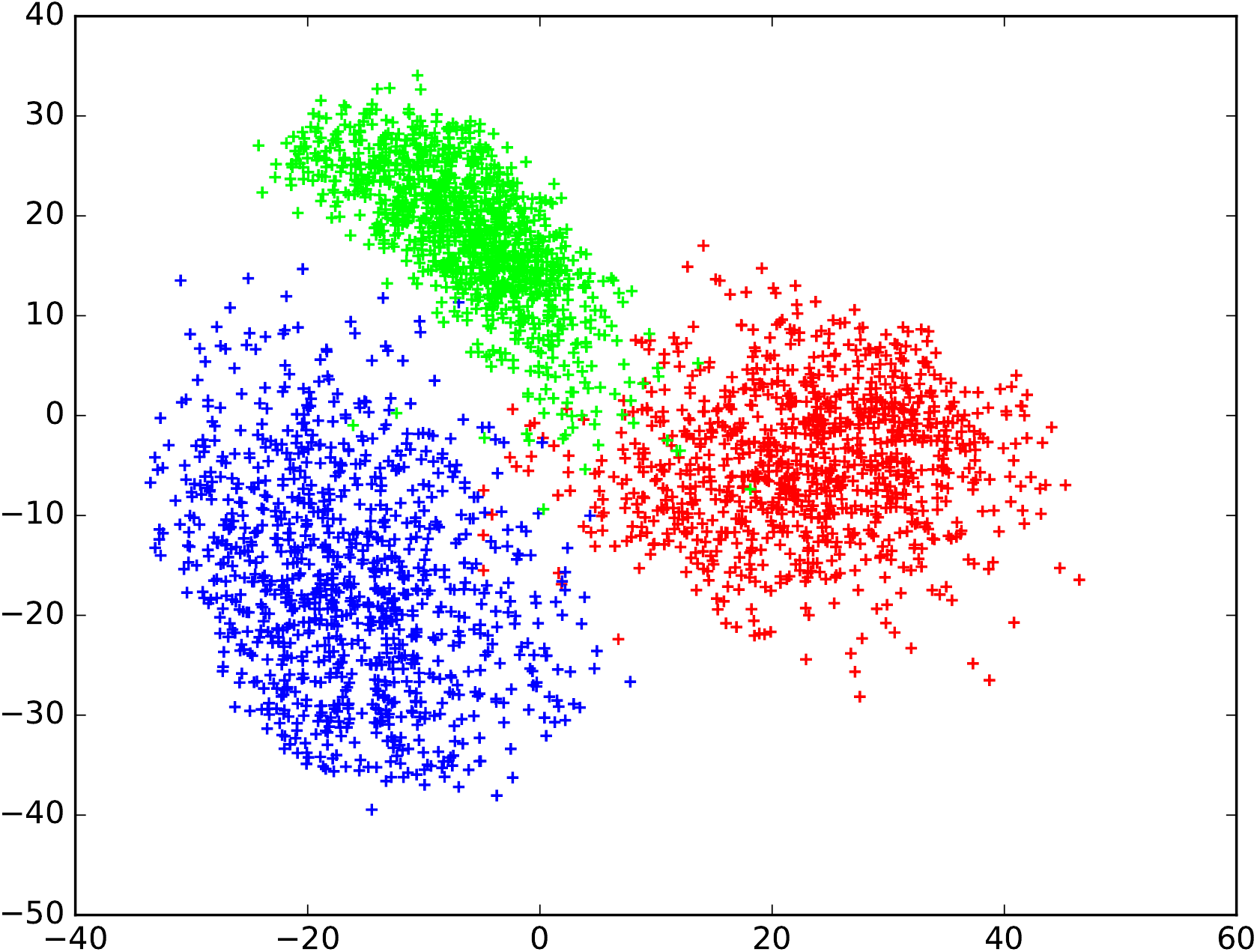}
  \caption{IPAE, $\beta=0.00001$, $N_j=32$\\ $err=0.80\pm0.24$.}
  \label{fig:2h}
\end{subfigure}%
\begin{subfigure}[b]{.33\textwidth}
\centering
  \includegraphics[height=38mm,width=.9\linewidth]{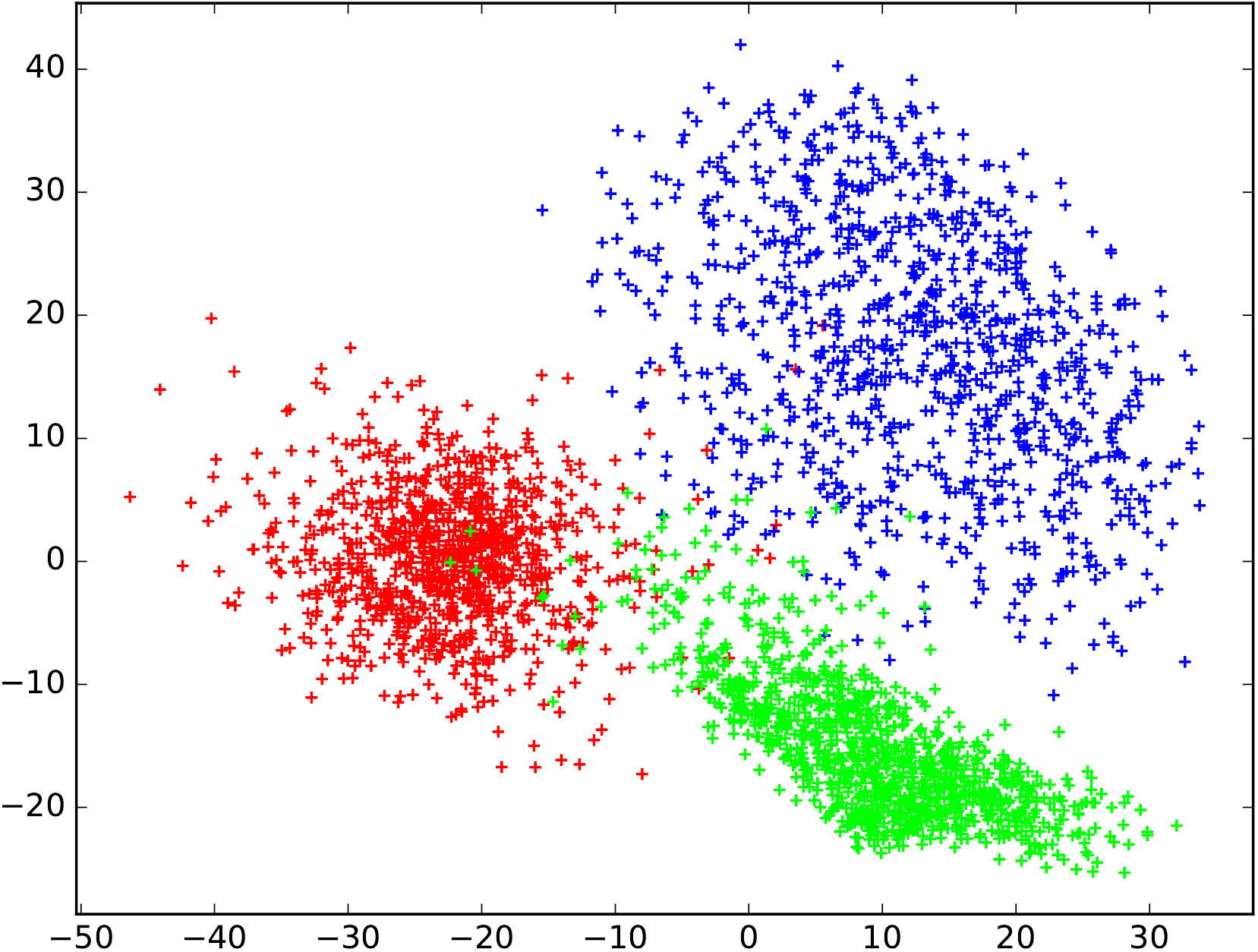}
  \caption{IPAE, $\beta=0.00001$, $N_j=64$\\ $err=0.74\pm0.21$.}
  \label{fig:2i}
\end{subfigure}%
\\

\caption{Visualization of encoding features $\mathbf{z}$ on test samples projected to $2$-dimensional spaces. In the second row, we use $N_j=1$.  We also train a linear SVM on the learned encoding features (without fine-tuning). The experiments are repeated $10$ times (including training of AEs), and average error $err$ (mean$\pm$std error (\%)) is given. The best two results are bolded. Note that, for $\beta\leq0.0001$, VAEs failed in training phases in some experiments due to computation of large $\sigma^2(\mathbf{z})$.}
\label{fig:mnist}
\end{figure}

In the next experiment, we take a subset of the MNIST to evaluate our method. We select $18000$ training samples belonging to three classes of digits $1$, $3$ and $4$. The network architecture, which was used to construct an encoder and a decoder, is as follows:
    \begin{equation*}
    \begin{split}
    \mathbf{x}\xrightarrow{\text{sigmoid}} 1024 \rightarrow
    \begin{cases}
        \mu(\mathbf{x})\\
        \sigma(\mathbf{x})\\
    \end{cases}
\dasharrow
\mathbf{z}\xrightarrow{\text{sigmoid}} 1024
\rightarrow 784
     \end{split}
     \label{eq:mnistipae}
    \end{equation*}
where $\mu(\mathbf{x})$ and $\sigma(\mathbf{x})$ are both $8$-dimensional. In this task, the distortion function $d(\cdot,\cdot)$ used in our objective \eqref{eq:obj} is
\begin{equation*}
\begin{split}
d(\mathbf{x},\tilde{\mathbf{x}})=-\sum \limits_{i=1}^M ( \mathbf{x}^{(i)}\log \tilde{\mathbf{x}}^{(i)} + (1-\mathbf{x}^{(i)}) \log (1-\tilde{\mathbf{x}}^{(i)}) )
\end{split}
\label{eq:bern}
\end{equation*}   
where ${\mathbf{x}^{(i)}}$ denotes the $i^{th}$ dimension of vector, and $\tilde{\mathbf{x}}$ is the reconstructed vector. This distortion used in VAE is interpreted as a Bernoulli distribution of $p(\mathbf{x}|\mathbf{z})$. The same learning procedure as that of the toy dataset is adopted. We visualize the distribution of encoding variable $\mathbf{z}$ of test samples by projecting it to $2$-dimensional space using PCA. 
% The choice of PCA for the projection method is to due to its linear mapping function to keep the Gaussian or non-Gaussian distribution of $p(\mathbf{z})$.
The results are shown in Figure \ref{fig:mnist}, together with test classification error (\%) obtained by training (without fine-tuning) a linear SVM on the learned encoding features. The experiments are repeated  $10$ times and average error is given (AEs are also re-trained at each time).

As we can see from the results, the proposed IPAEs provide better classification performance than VAEs. The best results obtained by IPAEs using $\beta=0.00001$, $N_j=8$ (see \ref{fig:2g}) outperform VAEs that use $\beta=0.001$ (see \ref{fig:2a}). Note that, VAEs which are trained using $\beta\leq0.0001$, failed in some experiments by obtaining too large $\sigma^2(\mathbf{z})$.
In terms of visualization,
% which suggests the proposed non-parametric method for estimation of $H(\mathbf{z})$ works better than 
the inter-class distance between samples tends to be larger, i.e. more linearly discriminative using smaller $\beta$ values, since the data is less compressed. Thus the classification error gets lower.
As we increase the values of $\beta$, the samples belonging to different categories are contracted, i.e. samples belonging to different categories are mapped to the same point (see e.g. the first and second row of the figure). It causes
the samples in the encoding space become less discriminative using both VAEs and IPAE. 
The reason can be that a larger compression degree causes some discriminative features to be removed. As a result, the classification error also increases.
Also, if we increase the $N_j$ value in the IPAE model in order to increase the interaction between more pairs of samples, 
then the samples belonging to the same category have a more compact distribution, meanwhile the error can be almost unchanged. Therefore, the inter-class discrimination is preserved (see \ref{fig:2g},\ref{fig:2h},\ref{fig:2i} compared with \ref{fig:2d}). 
Without employment of pair-wise interaction between samples, VAEs obtain a larger error if the compression degree is increased.

\section{Conclusions}

% In order to estimate mutual information, we propose a non-parametric method 

In this paper, we introduce an information theoretic framework for training of Auto-Encoders. In order to estimate mutual information, we
propose a non-parametric method that estimates the entropy of encoding variable $H(\mathbf{z})$.
We also give an information theoretic view of VAEs, which suggests that VAEs can be seen as parametric methods that estimate entropy. Experimental results show that the proposed IPAEs have more degree of freedom in terms of representation learning of features drawn from complex distributions such as Mixture of Gaussians, compared to VAEs. In our future work, we will consider application of our proposed non-parametric information regularization method to other type of generative models such as generative adversarial networks \cite{gan}.

% \begingroup
% \renewcommand\refname{}
% \input{main.bbl}
% \endgroup

\bibliographystyle{plain}
\bibliography{ref.bbl}
\end{document}